# TALISMAN: Targeted Active Learning for Object Detection with Rare Classes and Slices using Submodular Mutual Information

Suraj Kothawade[1], Saikat Ghosh[1], Sumit Shekhar[2], Yu Xiang[1], and Rishabh Iyer[1]

[1] University of Texas at Dallas
{suraj.kothawade, saikat.ghosh, yu.xiang, rishabh.iyer}@utdallas.edu
[2] Adobe Research
sushekha@adobe.com

**Abstract.** Deep neural networks based object detectors have shown great success in a variety of domains like autonomous vehicles, biomedical imaging, *etc.*, however their success depends on the availability of a large amount of data from the domain of interest. While deep models perform well in terms of overall accuracy, they often struggle in performance on rare yet critical data slices. For *e.g.*, detecting objects in rare data slices like "motorcycles at night" or "bicycles at night" for self-driving applications. Active learning (AL) is a paradigm to incrementally and adaptively build training datasets with a human in the loop. However, current AL based acquisition functions are not well-equipped to mine rare slices of data from large real-world datasets, since they are based on uncertainty scores or global descriptors of the image. We propose Talisman, a novel framework for **T**argeted **A**ctive **L**earning for object detect**I**on with rare slices using **S**ubmodular **M**utu**A**l i**N**formation. Our method uses the submodular mutual information functions instantiated using features of the region of interest (RoI) to efficiently target and acquire images with rare slices. We evaluate our framework on the standard PASCAL VOC07+12 [8] and BDD100K [31], a real-world large-scale driving dataset. We observe that Talisman consistently outperforms a wide range of AL methods by $\approx 5\% - 14\%$ in terms of average precision on rare slices, and $\approx 2\% - 4\%$ in terms of mAP. The code for Talisman is available here: https://github.com/surajkothawade/talisman.

## 1 Introduction

Deep learning approaches for object detection have made a lot of progress, with accuracies improving consistently over the years. As a result, object detection technology is extensively being used and deployed in applications like self-driving cars and medical imaging, and is approaching human performance. One critical aspect, though in high-stake applications like self-driving cars and medical imaging, is that the cost of failure is very high. Even a single mistake in the detection and specifically a false-negative (*e.g.*, missing a pedestrian on a highway or a motorcycle at night) can result in a major and potentially fatal accident[3].

An important aspect in such problems is that there are a number of rare yet critical slices of objects and scenarios. Because many of these rare slices are severely under-represented in the data, deep learning based object detectors often perform poorly in such scenarios. Some examples of such data slices are "motorcycles at night", "pedestrians on a highway", and "bicycles at night". Fig. 1 shows the distribution of slices in the BDD100K [31] dataset. As is evident, these slices are very rare – for instance, the number of motorcycles at night, is 0.094% of the number of cars in the dataset.

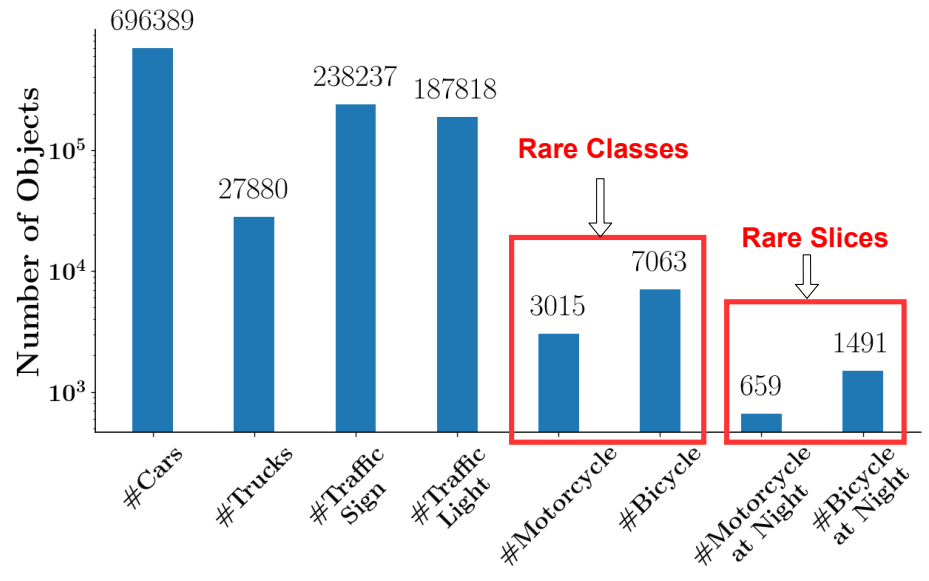

Fig. 1: Problem Statement: Rare classes and Rare slices in BDD100K [31]. Motorcycle and bicycle classes have the least number of objects, thereby making them *rare classes*, on which the model performs the worst in terms of average precision (AP). Further, motorcycle/bicycle objects at night are *rarer*, thereby making them *rare slices* on which the model performs the worst.

This causes a more pronounced issue in the limited data setting. To understand the effect of this imbalance, we trained a Faster-RCNN Model [23] on a small subset of BDD100K (5% of the dataset) and we noticed a significant difference in mAP between "cars" class (around 55% mAP) and "motorcycle" (around 9% mAP). This gap is even more pronounced for rare slices. Active learning based data sampling is an increasingly popular paradigm for training deep learning classifiers and object detectors [6,25,29,10] because such approaches significantly reduce the amount of labeled data required to achieve a certain desired accuracy. On the other hand, current active learning based paradigms are heavily dependent on aspects like uncertainty and diversity, and often miss rare slices of data. This is because such slices, though critical for the end task, are a small fraction of the full dataset, and play a negligible role in the overall accuracy.

## 1.1 Our Contributions

In this paper, we propose TALISMAN, a novel active learning framework for object detection, which (a) provides a mechanism to encode the similarity between an unlabeled image and a small query set of targeted examples (e.g., images with "motorcycles at night" RoIs), and (b) mines these examples in a scalable manner from a large unlabeled set using the recently proposed submodular mutual information functions. We also provide an approach where we can mine examples based on multiple such rare slices. Similar to standard active learning, TALISMAN is an interactive human-in-the-loop approach where images are chosen iteratively and provided to a human for labeling. However, the key difference is that TALISMAN does the selection by targeting rare slices using only a few exemplars. The overview of targeted selection using TALISMAN is shown in Fig. 3.

[3] An unfortunate example of this is the self-driving car crash with Uber: `https://www.theverge.com/2019/11/6/20951385/uber-self-driving-crash-death-reason-ntsb-dcouments` where the self-driving car did not detect a pedestrian on a highway at night, resulting in a fatal accident.

Empirically, we demonstrate the utility of TALISMAN on a diverse set of rare slices that occur in the real-world. Specifically, we see that TALISMAN outperforms the best baseline by significant margins on different rare slices (c.f. Fig. 2).

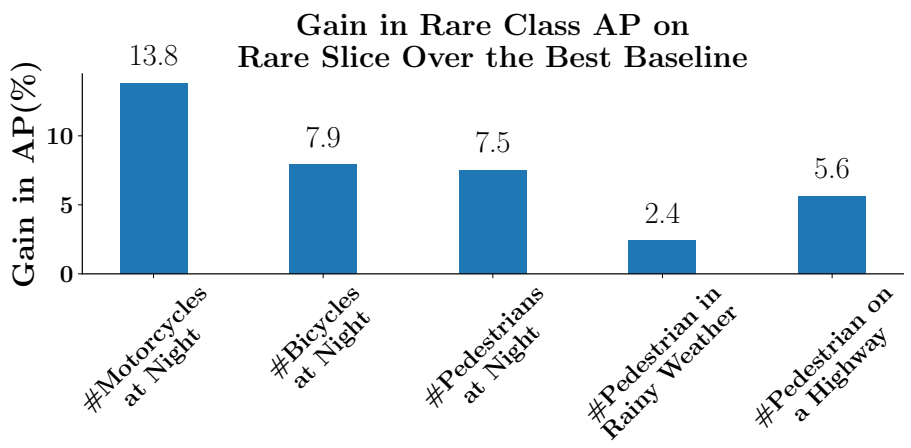

Fig. 2: Efficiency of TALISMAN over the best-performing baseline on a variety of rare slices in BDD100K

## 2 Related Work

A number of recent works have studied deep active learning for image classification [27,2,30,26,16,19,5]. The most common approach for active learning is to select the most uncertain examples. These include approaches like ENTROPY [27], LEAST CONFIDENCE [29], and MARGIN [24]. One challenge of this approach is that all the samples within a batch can be potentially similar, even though they are uncertain. Hence, a number of recent works have ensured that we select examples that are both uncertain and diverse. Examples include BADGE [2], FASS [30], BATCH-BALD [18], CORESET [26], and so on.

Recently, researchers have started applying active learning to the problem of object detection. [6] proposed an uncertainty sampling based approach for active object detection, while [25] proposed a 'query-by-committee' paradigm to select the most uncertain items for object detection. Recently [10], studied several scoring functions for active learning, including entropy based functions, coreset based functions, and so on. [15] proposed an active learning approach based on the localization of the detections, and studied the role of two metrics called "localization tightness" and "localization stability" as uncertainty measures. [7] studied active learning in the setting of users providing weak supervision (i.e., just suggesting the label and a rough location as opposed to drawing bounding boxes around the objects). All these approaches have shown significant labeling cost reductions and gains in accuracy compared to random sampling. However, the major limitation with these approaches (which are mostly variations of uncertainty) is that they focus on the overall accuracy, and do not necessarily try to select instances specific to certain rare yet critical data slices. To overcome these limitations, we provide a generalized paradigm for active learning in object detection, where we can target specific rare data slices.

A related thread of research is the use of the recently proposed submodular information measures [13] for data selection and active learning. [17] extended the work of [13] and proposed a general family of *parameterized submodular information measures* for guided summarization and data subset selection. [19] use the submodular information measures for active learning in the image classification setting to address realistic scenarios like imbalance, redundancy, and out-of-distribution data. Finally, [20] use the submodular information measures for personalized speech recognition. To our knowledge, this is the first work which proposes an active learning framework for object detection capable of handling rare slices of data.

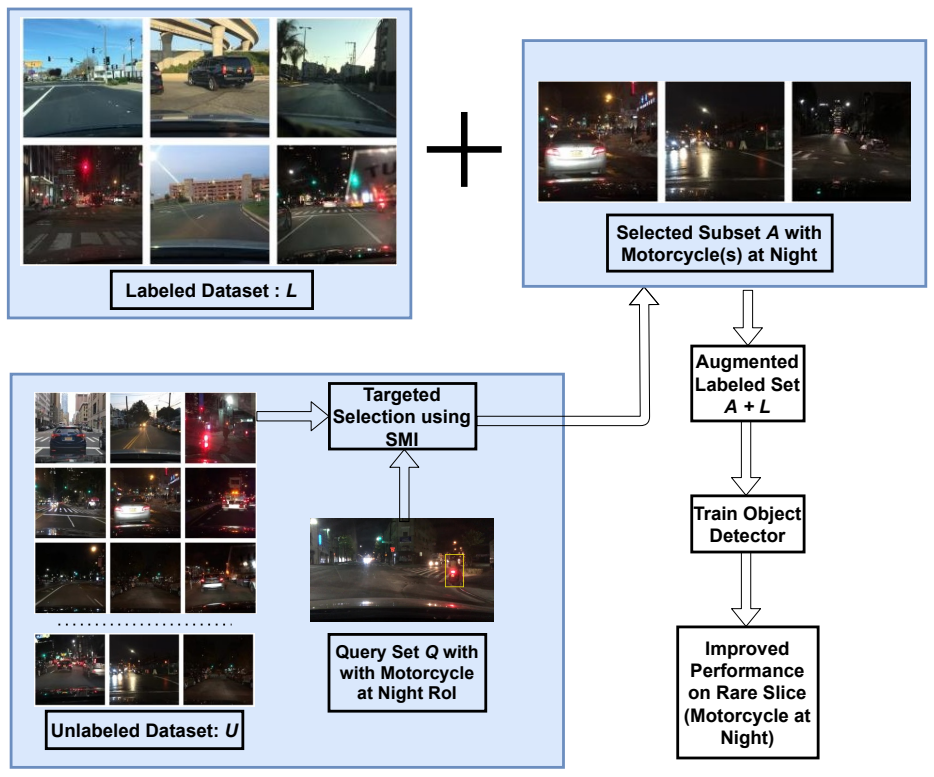

Fig. 3: Targeted Selection using TALISMAN for one round of targeted active learning. Motorcycles at night is a rare slice in the labeled data. We mine images from the unlabeled set that semantically similar to the RoIs in the query set by using the submodular mutual information (SMI) functions. These images are then labeled and added to the labeled data to improve performance on the rare slice.

## 3 Background

In this section, we discuss different submodular functions and their mutual information instantiations.

### 3.1 Submodular Functions

Submodular functions are an appealing class of functions for data subset selection in real-world applications due to their diminishing returns property and their ability to model properties of a good subset, such as diversity, representation and coverage [28,3,4,14]. Consider an unlabeled set $\mathcal{U} = \{1, 2, 3, \cdots, n\}$ and a set function $f : 2^{\mathcal{U}} \to \mathbb{R}$. Formally, $f$ is defined to be submodular [9] if for $x \in \mathcal{U}$, $f(\mathcal{A} \cup x) - f(\mathcal{A}) \geq f(\mathcal{B} \cup x) - f(\mathcal{B})$, $\forall \mathcal{A} \subseteq \mathcal{B} \subseteq \mathcal{U}$ and $x \notin \mathcal{B}$. For data subset selection and active learning, a number of recent approaches [30,16] use $f$ as an acquisition function to obtain a real-valued score for $f(\mathcal{A})$, where $\mathcal{A} \subseteq \mathcal{U}$. Given a budget $\mathcal{B}$ (the number of elements to select at every round of subset selection of batch active learning), the optimization problem is: $\max_{\mathcal{A}:|\mathcal{A}| \leq \mathcal{B}} f(\mathcal{A})$. Two examples of submodular functions that we use in this work are Facility Location (FL) and Graph Cut (GC) functions (see Tab. 1(a)). They are instantiated by using a similarity matrix $S$, that stores the similarity scores $S_{ij}$ between any two data points $i, j$. The submodular functions admit a constant factor approximation $1 - \frac{1}{e}$ [22] for cardinality constraint maximization. Importantly, submodular maximization can be done in *near-linear time* using variants of greedy algorithms [21].

### 3.2 Submodular Mutual Information (SMI)

While submodular functions are a good choice of functions for standard active learning, in this work, we want to not only select the most informative and diverse set of points, but also select points which are *similar* to a specific target slice (typically only a few examples from a rare slice). The Submodular mutual information (SMI) functions capture this second property and are defined as $I_f(\mathcal{A}; \mathcal{Q}) = f(\mathcal{A}) + f(\mathcal{Q}) - f(\mathcal{A} \cup \mathcal{Q})$, where $\mathcal{Q}$ is a query or target set (e.g., a few sample images of "motorcycles at night"). Intuitively, maximizing the SMI functions ensure that we obtain *diverse* subsets that are *relevant* to a query set $\mathcal{Q}$. We discuss the details of the SMI functions used in our work in the next section.

### 3.3 Specific SMI Functions Used In TALISMAN

We adapt the mutual information variants of Facility Location (FL) and Graph Cut (GC) functions [17] for targeted active learning.

**Facility Location:** The FL function models representation (i.e., it picks the most representative points or "centroids"). The FL based SMI function called FLMI can be written as $I_f(\mathcal{A}, \mathcal{Q}) = \sum_{i \in \mathcal{Q}} \max_{j \in \mathcal{A}} S_{ij} + \sum_{i \in \mathcal{A}} \max_{j \in \mathcal{Q}} S_{ij}$ [17]. This function models representation as well as query relevance.

**Graph Cut:** The GC function models diversity and representation, and has modeling properties similar to FL. The SMI variant of GC is defined as GCMI, which maximizes the pairwise similarity between the query set and the unlabeled set. The GCMI function can be written as $I_f(\mathcal{A}; \mathcal{Q}) = 2 \sum_{i \in \mathcal{A}} \sum_{j \in \mathcal{Q}} S_{ij}$.

Table 1: Instantiations of different submodular functions.

(a) Instantiations of Submodular functions.

| SF | $f(\mathcal{A})$ |
|---|---|
| FL | $\sum_{i \in \mathcal{U}} \max_{j \in \mathcal{A}} S_{ij}$ |
| GC | $\sum_{i \in \mathcal{A}, j \in \mathcal{U}} S_{ij} - \sum_{i,j \in \mathcal{A}} S_{ij}$ |

(b) Instantiations of SMI functions.

| SMI | $I_f(\mathcal{A}; \mathcal{Q})$ |
|---|---|
| FLMI | $\sum_{i \in \mathcal{Q}} \max_{j \in \mathcal{A}} S_{ij} + \sum_{i \in \mathcal{A}} \max_{j \in \mathcal{Q}} S_{ij}$ |
| GCMI | $2 \sum_{i \in \mathcal{A}} \sum_{j \in \mathcal{Q}} S_{ij}$ |

Tab. 1(a) and (b) demonstrate the SMI functions we will use in this work and the corresponding submodular functions instantiating them. Note that in [13,17], a number of other SMI functions and instantiations have been proposed. However, keeping scalability to large datasets in mind (see Sec. 4.4), we only focus on these two.

# 4 TALISMAN: Our Targeted Active Learning Framework for Object Detection

## 4.1 TALISMAN Framework

In this section, we present TALISMAN, our targeted active learning framework for object detection. We show that TALISMAN can efficiently target any imbalanced scenario with rare classes or rare slices. We summarize our method in Algorithm 1, and illustrate it in Fig. 4. The core idea of our framework lies within instantiating the SMI functions such that they can mine for images from the unlabeled set which contain proposals semantically similar to the region of interests (RoIs) in the query set. The query set contains exemplars of the rare slice that we want to target.

**Algorithm 1** TALISMAN: Targeted AL Framework for Object Detection (Illustration in Fig. 4)

**Require:** Initial labeled set of data points: $\mathcal{L}$, large unlabeled dataset: $\mathcal{U}$, small query set $\mathcal{Q}$, object detection model $\mathcal{M}$, batch size: $B$, number of selection rounds: $N$.

**for** selection round $i = 1 : N$ **do**
  Train model $\mathcal{M}$ on the current labeled set $\mathcal{L}$ and obtain parameters $\theta_i$
  Compute $S \in \mathbb{R}^{|\mathcal{Q}| \times |\mathcal{U}|}$ such that: $S_{qu} \leftarrow \textsc{TargetedSim}(\mathcal{M}_{\theta_i}, \mathcal{I}_q, \mathcal{I}_u)$, $\forall q \in \mathcal{Q}, \forall u \in \mathcal{U}$ {Algorithm 2}
  Instantiate a submodular function $f$ based on $S$.
  $\mathcal{A}_i \leftarrow \text{argmax}_{\mathcal{A} \subseteq \mathcal{U}, |\mathcal{A}| \leq B} I_f(\mathcal{A}; \mathcal{Q})$ {Greedy maximization of SMI function to select a subset $\mathcal{A}$}
  Get labels $L(\mathcal{A}_i)$ for batch $\mathcal{A}_i$ and $\mathcal{L} \leftarrow \mathcal{L} \cup L(\mathcal{A}_i)$, $\mathcal{U} \leftarrow \mathcal{U} - \mathcal{A}_i$
**end for**
**Return** trained model $\mathcal{M}$ and parameters $\theta$.

We start with training an object detection model $\mathcal{M}$ on an initial labeled set $\mathcal{L}$. Using $\mathcal{M}$, we compute embeddings of the query set $\mathcal{Q}$ and the unlabeled set $\mathcal{U}$. Next, we compute pairwise cosine similarity scores $S_{qu}, \forall q \in \mathcal{Q}, \forall u \in \mathcal{U}$ to obtain a similarity matrix $S \in \mathbb{R}^{|\mathcal{Q}| \times |\mathcal{U}|}$. We discuss the details of computing $S_{qu}$ for a single query image $q$ and a single unlabeled image $u$ in Sec. 4.2. Using the similarity matrix $S$, we instantiate the SMI function $I_f(\mathcal{A}; \mathcal{Q})$ as discussed in Sec. 3 (note that both the SMI functions we consider in this work are similarity based functions). Finally, we acquire a subset $\mathcal{A}$ that contains regions that are semantically similar to the RoI in $\mathcal{Q}$ by maximizing the SMI function $I_f(\mathcal{A}; \mathcal{Q})$:

$$\max_{\mathcal{A} \subseteq \mathcal{U}, |\mathcal{A}| \leq B} I_f(\mathcal{A}; \mathcal{Q}). \tag{1}$$

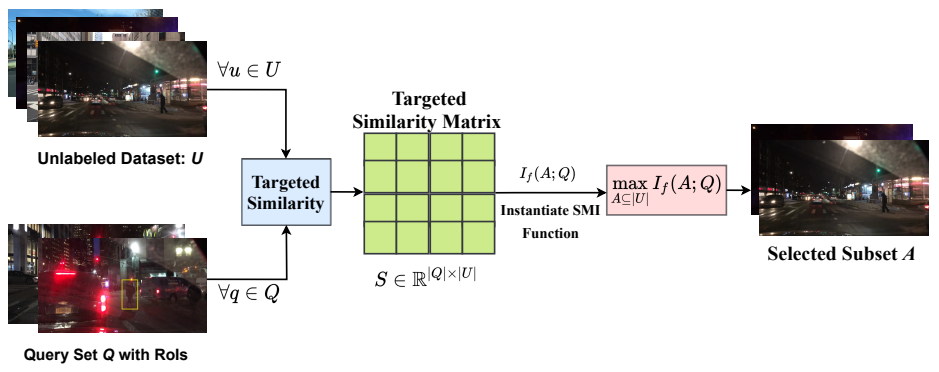

Fig. 4: Architecture of TALISMAN during one round of targeted active learning. We illustrate the targeted similarity computation in Fig. 5.

Since this function is submodular (i.e. $I_f(\mathcal{A}; \mathcal{Q})$ is submodular in $\mathcal{A}$ for a fixed query set $\mathcal{Q}$), we use a greedy algorithm [22] (to solve Equ. (1) and Line 5 in Algorithm 1) which ensures a $1 - \frac{1}{e}$ approximation guarantee of the optimal solution.

### 4.2 Targeted Similarity Computation

We summarize our method for targeted similarity computation in Algorithm 2 and illustrate it in Fig. 5. For simplicity, consider a single query image $\mathcal{I}_q \in \mathcal{Q}$ with $T$ RoIs (targets) indicating a rare slice, and an unlabeled image $\mathcal{I}_u \in \mathcal{U}$ with $P$ region proposals obtained using a region proposal network (RPN). Using $\mathcal{M}$ that is trained on $\mathcal{L}$, we compute the embedding of the RoIs in $\mathcal{I}_q$ to obtain $\mathcal{E}_q \in \mathbb{R}^{T \times D}$, and for the proposals of $\mathcal{I}_u$ to obtain $\mathcal{E}_u \in \mathbb{R}^{P \times D}$. Here, $D$ denotes the dimensionality of each feature vector representing a RoI or region proposal. We use the embeddings $\mathcal{E}_q$ and $\mathcal{E}_u$ to represent $\mathcal{I}_q$ and $\mathcal{I}_u$ respectively. We use these embeddings to compute the targeted similarity (see Algorithm 2).

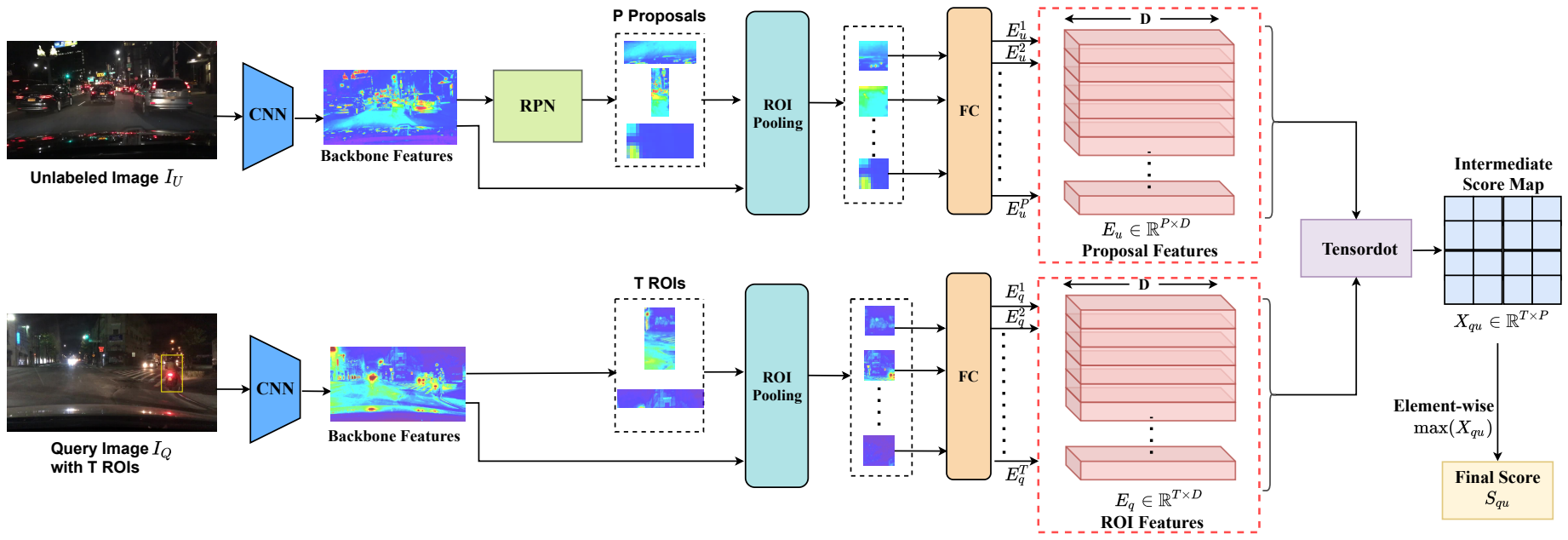

Fig. 5: TargetedSim: Targeted Similarity computation in Talisman.

In order to compute cosine similarity between $\mathcal{E}_q$ and $\mathcal{E}_u$ efficiently, we L2-normalize along the feature dimension of length $D$. This enables us to highly parallelize the similarity computation via off-the-shelf GPU enabled dot product [4] implementations. Next, we compute the dot product along the feature dimension to obtain pairwise similarities between $T$ RoIs in $\mathcal{I}_q$ and the $P$ proposals in $\mathcal{I}_u$ which gives us RoI-proposal score map $\mathcal{X}_{qu} \in \mathbb{R}^{T \times P}$. Finally, we assign the similarity score $S_{qu}$ between $\mathcal{I}_q$ and $\mathcal{I}_u$ by computing the *element-wise* maximum of $\mathcal{X}_{qu}$, which entails the best matching proposal of the $P$ region proposals to some query RoI in the $T$ RoIs.

[4] See torch.tensordot

**Algorithm 2** TargetedSim: Targeted Similarity Matching (Illustration in Fig. 5)

**Require:** Local feature extraction model $F_\theta$, $\mathcal{I}_q \in \mathcal{Q}$ with $T$ RoIs and $\mathcal{I}_u \in \mathcal{U}$ with $P$ region proposals.
1: $\mathcal{E}_q \leftarrow F_\theta(\mathcal{I}_q)$ $\{\mathcal{E}_q \in \mathbb{R}^{T \times D}\}$
2: $\mathcal{E}_u \leftarrow F_\theta(\mathcal{I}_u)$ $\{\mathcal{E}_u \in \mathbb{R}^{P \times D}\}$
3: $\mathcal{X}_{qu} \leftarrow$ Cosine_Similarity$(\mathcal{E}_q, \mathcal{E}_u)$ $\{\mathcal{X}_{qu} \in \mathbb{R}^{T \times P}$. Compute Cosine similarity along the feature dimension$\}$
4: $S_{qu} \leftarrow \max(\mathcal{X}_{qu})$ {Element-wise Max, $S_{qu}$ represents the score between the best matching proposal $j \in P$ to some query RoI $i \in T$}
5: **Return** Similarity score $S_{qu}$

### 4.3 Using TALISMAN to Mine Rare Slices

A critical input to Talisman (Algorithm 1) is the query set $\mathcal{Q}$. The query set consists of a specific target slice, which could be a rare class (e.g. "motorcycles") or a rare slice ("motorcycles at night"). In our experiments, we study the role of Talisman for both scenarios. For our setting to be realistic, we need to ensure that $\mathcal{Q}$ is tiny – since these are rare slices, we cannot assume that we have access to numerous of these rare examples. For this reason, we set $\mathcal{Q}$ to be between 2 and 5 examples in our experiments. It is worth noting that since the SMI functions naturally model relevance to the query set and diversity within the selected subset, they pick a diverse set of data points which are relevant to the query set $\mathcal{Q}$.

### 4.4 Scalability of TALISMAN

A key factor in the efficiency of Talisman is the choice of SMI functions Flmi and Gcmi. The memory and time complexity of computing the similarity kernel for both these functions is only $|\mathcal{Q}| \times |\mathcal{U}|$ – since $\mathcal{Q}$ is a tiny held-out set of the examples from the rare slice (of size 2 to 5), the time complexity of creating and storing the Flmi and Gcmi functions is only $O(\mathcal{U})$. For the greedy algorithm [22], we use memoization [12]. This ensures that the complexity of computing the gains for both Flmi and Gcmi functions is in fact $O(|\mathcal{Q}|$, which is a constant, so the amortized complexity using the lazy greedy algorithm is $|\mathcal{U}| \log |\mathcal{U}|$. We can also use the lazier than lazy greedy algorithm [21], which ensures that the worst case complexity of the greedy algorithm is only $|\mathcal{U}|$. As a result, both Flmi and Gcmi can be optimized in linear time (with respect to the size of the unlabeled set), thereby ensuring that Talisman can scale to very large datasets.

## 5 Experimental Results

In this section, we empirically evaluate the effectiveness of Talisman for a wide range of real-world scenarios where the dataset has one or more rare classes or rare slices. We do so by comparing the performance of Talisman instantiated SMI functions (Tab. 1(b)) with existing active learning approaches using a wide variety of metrics, namely the mean average precision (mAP), average precision (AP) of the rare slice, and the number of data points selected that belong to the rare slice. We summarize all notations used in Appendix. A.

### 5.1 Experimental Setup

We apply TALISMAN for object detection tasks on two diverse public datasets: 1) the standard PASCAL VOC07+12 (VOC07+12) [8] and 2) BDD100K [31], a large scale driving dataset. VOC07+12 has 16,551 images in the training set and 4,952 images in the test set which come from the test set of VOC07. BDD100K consists of 70K images in the training set, 10K images in the validation set and 20K images in the test set. Since the labels for the test set are not publicly available, we use the validation set for evaluation. For active learning (AL), we split the training set into the labeled set $\mathcal{L}$ and unlabeled set $\mathcal{U}$.

Since the problem of targeted active learning is more about sampling *objects* semantically similar to a region of interest, we create the initial seed set for AL by randomly sampling images such that an *object-level* budget is satisfied for each class. This allows us to simulate multiple scenarios with rare classes or rare slices. In the following sections, we provide the individual splits for $\mathcal{L}$,$\mathcal{U}$, and $\mathcal{Q}$ in each rare class or rare slice scenario.

In all the AL experiments discussed below, we use a common training procedure and hyperparameters to ensure fair comparison across all acquisition functions. We use standard data augmentation techniques like random flips followed by normalization. For all experiments on both datasets, we train a Faster RCNN model [23] based on a ResNet50 backbone [11]. In each round of AL, we reinitialize the model parameters, and train the model for 150 epochs using SGD with momentum. The initial learning rate is set to 0.001 with a step size of 3, the momentum and weight decay are set to 0.9 and 0.0005 respectively. For comparing multiple acquisition functions in the AL loop, we start with an identical model that is trained on the initial labeled set $\mathcal{L}$. All the experiments were run 5× on a V100 GPU and the error bars (std deviation) are reported.

### 5.2 Baselines in All Scenarios

We compare TALISMAN instantiated SMI functions with multiple AL baselines: namely ENTROPY [27,10], Targeted Entropy (T-ENTROPY), Least Confidence (LEAST-CONF) [29], MARGIN [24], FASS [30], CORESET [26], BADGE [2] and RANDOM sampling. Below, we discuss the details for each baseline:

**Entropy [27,10]:** We compute the entropy for each region proposal of a specific class by using the probability scores generated by the model $\mathcal{M}$. This entropy is computed as follows:

$$\mathcal{H}(R_c) = -R_c \log R_c - (1 - R_c) \log(1 - R_c), \tag{2}$$

where $R_c$ represents the probability for class $c$ at the region proposal $R$. We set the number of region proposals $P = 300$ in our experiments. We compute the final entropy score $s$ of an unlabeled image $\mathcal{I}_u$ by taking the maximum across all $C$ classes for each proposal followed by an average across all proposals as follows:

$$s = \underset{R}{avg} \max_{c \in C} \mathcal{H}(R_c). \tag{3}$$

**Targeted Entropy:** In order to encourage Entropy sampling to select more points relevant to the query set, we make Entropy *target-aware* by following a two-step process. First, we select the top-$K$ data points with maximum entropy. Next, we compute a $|\mathcal{Q}| \times K$ similarity matrix using these top-$K$ uncertain data points and the query set $\mathcal{Q}$ (as done in lines 3-5 of Algorithm 1). Finally, we select the top-$B$ samples from these top-$K$ samples that have a region semantically similar to some RoI in $\mathcal{Q}$. We refer to this method as Targeted Entropy (T-Entropy). We set $K > B$ in our experiments so that T-Entropy has enough samples to choose from.
**Least Confidence [29]:** Least confidence (Least-Conf) is another intuitive acquisition function based on uncertainty. We compute the Least-Conf score $s$ of each data point by averaging over the minimum predicted class probability of $P$ region proposals as follows:

$$s = \underset{R}{avg} \min_{c \in C} R_c \tag{4}$$

Intuitively, we select the bottom $B$ data points that have the smallest predicted class probability scored by $s$.
**Margin [24]:** For Margin sampling, we score each data point by averaging over the difference between the top two predicted class probabilities of $P$ region proposals as follows:

$$s = \underset{R}{avg} \min_{c1, c2 \in C} R_{c1} - R_{c2} \tag{5}$$

We use the score $s$ to select $B$ data points that have the least difference in the probability score of the first and the second most probable labels.
**FASS [30]:** In order to encourage uncertainty and diversity using Fass, we first select top $K \times B$ uncertain data points using Entropy. Finally, we select top-$B$ data points using the facility location (FL) submodular function. We use FL since it performs the best in [30].
**Coreset [26]:** Coreset is a diversity based approach that selects core-sets such that the geometric arrangement of the superset is maintained. The core-sets are acquired using a greedy $k$-center clustering approach. In our experiments, we use the features of the last convolutional layer to represent each data point.
**Badge [2]:** Badge proposes to select diverse and uncertain data points that have a high gradient magnitude. The gradients are computed using hypothesized labels and distanced from previously selected data points using k-means++ [1]. In our experiments, we use the gradients from the penultimate convolutional layer of the ResNet50 backbone of the Faster RCNN model.
**Random:** For Random, we select $B$ data points randomly.

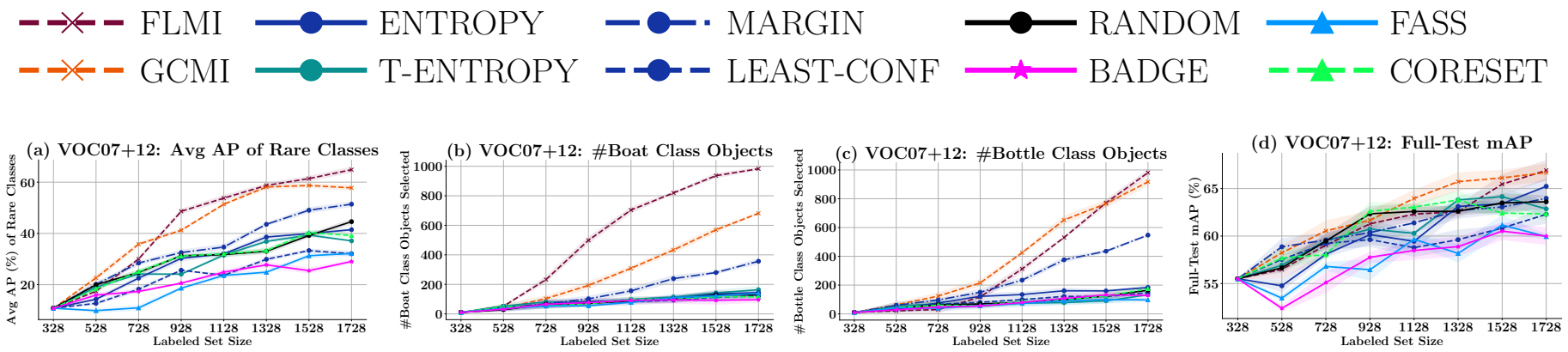

Fig. 6: Active Learning with rare classes on VOC07+12. Plot (a) shows the average AP of the rare classes, plots (b-c) show the number of boat and bottle objects selected respectively, plot (d) shows the mAP on the VOC07+12 test set. We observe that the SMI functions (Flmi, Gcmi) outperform other baselines by $\approx 8\% - 10\%$ average AP of the rare classes.

### 5.3 Rare Classes

**Dataset setting:** We conduct the experiments for the rare classes scenario on the VOC07+12 dataset. In particular, we create the initial labeled set, $\mathcal{L}$ which simulates the rare classes by creating a class imbalance at an object level. Let $\mathcal{C}_i^{\mathcal{L}}$ be the number of objects from a rare (infrequent) class $i$ and $\mathcal{B}_j^{\mathcal{L}}$ be the number of objects from a frequent class $j$. The initial labeled set $\mathcal{L}$ is created such that the imbalance ratio between $\mathcal{C}_i^{\mathcal{L}}$ and $\mathcal{B}_j^{\mathcal{L}}$ is *at least* $\rho$, *i.e.*, $\rho \leq (\mathcal{B}_j^{\mathcal{L}}/\mathcal{C}_i^{\mathcal{L}})$. All the remaining data points are used in the unlabeled set $\mathcal{U}$. In our experiments, we choose two classes to be rare from VOC07+12: 'boat' and 'bottle'. We do so due to two reasons: 1) they are by default the most uncommon objects in VOC, thereby making them the natural choice, and 2) they are comparatively smaller objects than other classes like 'sofa', 'chair', 'train', *etc.* We use a small query set $\mathcal{Q}$ containing 5 randomly chosen data points representing the rare classes (RoIs). We construct the initial labeled set by setting $\rho = 10$, $|\mathcal{C}^{\mathcal{L}}| = 20$ and $|\mathcal{B}^{\mathcal{L}}| = 2858$. This gives us an initial labeled seed set of size $|\mathcal{L}| = 1143$ images. Note that the imbalance ratio is not exact because objects of some classes are predominantly present in most images, thereby increasing the size of $|\mathcal{B}^{\mathcal{L}}|$.

**Results:** In Fig. 6, we compare the performance of Talisman on the rare classes scenario in VOC07+12 [8]. We observe that Talisman significantly outperforms all state-of-the art uncertainty based methods (Entropy, Least-Conf, and Margin) by $\approx 8\% - 10\%$ (Fig. 6(a)) in terms of average precision (AP) on the rare classes and by $\approx 2\% - 3\%$ in terms of mAP (Fig. 6(d)). This improvement is performance is because the Talisman instantiated functions (Gcmi and Flqmi) are able to select more data points that contain regions with objects belonging to the rare classes (see Fig. 6(c)). Interestingly, the Talisman functions were also able to give a fair treatment to multiple rare classes at the same time by selecting significant number of objects belonging to both the rare classes ('boat' and 'bottle', see Fig. 6(b,c)). This suggests that Talisman is able to select diverse data points by appropriately targeting regions containing rare class objects in the query image.

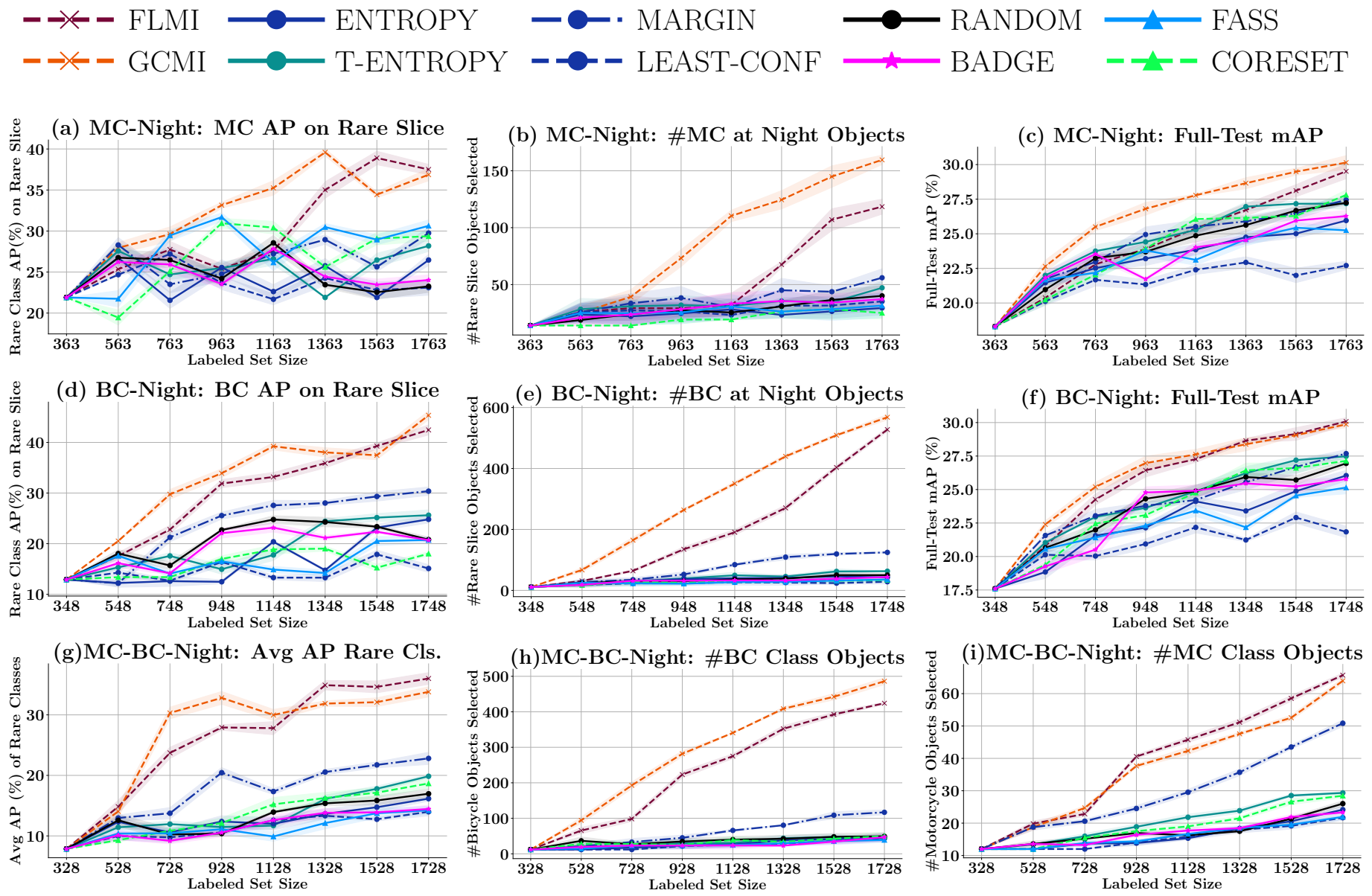

Fig. 7: Active Learning with Motorcycle (**MC**) at Night (top row) and Bicycle (**BC**) at Night (bottom row) rare slices on BDD100K. Left side plots (a,d,g) show the AP of the rare class on the rare slice of data, center plots (b,e) show the number of objects selected that belong to the rare slice, and right side plots (c,f) show the mAP on the full test set of BDD100K. We observe that the SMI functions (Flmi, Gcmi) outperform other baselines by $\approx 5\% - 14\%$ AP of the rare class on the rare slice. In (h,i), we show that Talisman selects more objects from multiple rare slices in comparison to the existing methods.

### 5.4 Rare Slices

**Dataset setting:** We chose BDD100K [31] since it is a realistic, large, and challenging dataset that allows us to evaluate the performance of Talisman on datasets with naturally occurring rare slices. Since we want to evaluate rare slices, the procedure to simulate the initial labeled set and the evaluation is slightly different from the rare classes experiment in the above section. In the following sections, we discuss experiments where the initial labeled set $\mathcal{L}$ has a rare slice made of a *class* and an *attribute*. For instance, *motorcyles* (class) at *night* (attribute), *pedestrians* (class) in *rainy weather* (attribute), *etc.* Let $|\mathcal{O}_c^A|$ be the number of objects in $\mathcal{L}$ that belong to class $c$ and attribute $A$. Concretely, we create a balanced initial labeled set $\mathcal{L}$, such that each class $c$ contributes an equal number of objects, *i.e.*$|\mathcal{O}_{c1}| = |\mathcal{O}_{c2}|, \forall c1, c2$. Let $i$ be the class involved in the rare slice. We simulate the rare slice by creating an imbalance in $\mathcal{O}_i$ based on an attribute $A$ such that the ratio between the number of objects of class $i$ with attribute $A$ and the ones without attribute $A$ (denoted by $\tilde{\mathbb{A}}$) is *at least* $\rho$, *i.e.*$\rho \leq (|\mathcal{O}_i^{\tilde{\mathbb{A}}}|/|\mathcal{O}_i^A|)$. In all the rare slice experiments, we start with an initial labeled set by setting $\rho = 10, |\mathcal{O}_i^{\tilde{\mathbb{A}}}| = 100$, and $|\mathcal{O}_i^{A}| = 10$. For all other classes $j$, we randomly pick objects such that $|\mathcal{O}_j| = 110$. Note that we use a small query set in all experiments ($\approx 3-5$ images). The exact number of images in $\mathcal{L}$ and $\mathcal{Q}$ for each experiment is given in Appendix. B. For evaluation, we compare the performance of TALISMAN using three metrics: 1) *Rare Class Rare Slice AP*: the average precision (AP) of the 'rare class' (*e.g.motorcycle*) on the 'rare slice' (*e.g.night*), 2) *# Rare Slice Objects*: the number of objects selected that belong to the rare slice, and 3) *Overall Test mAP*: the mAP on the complete test set.

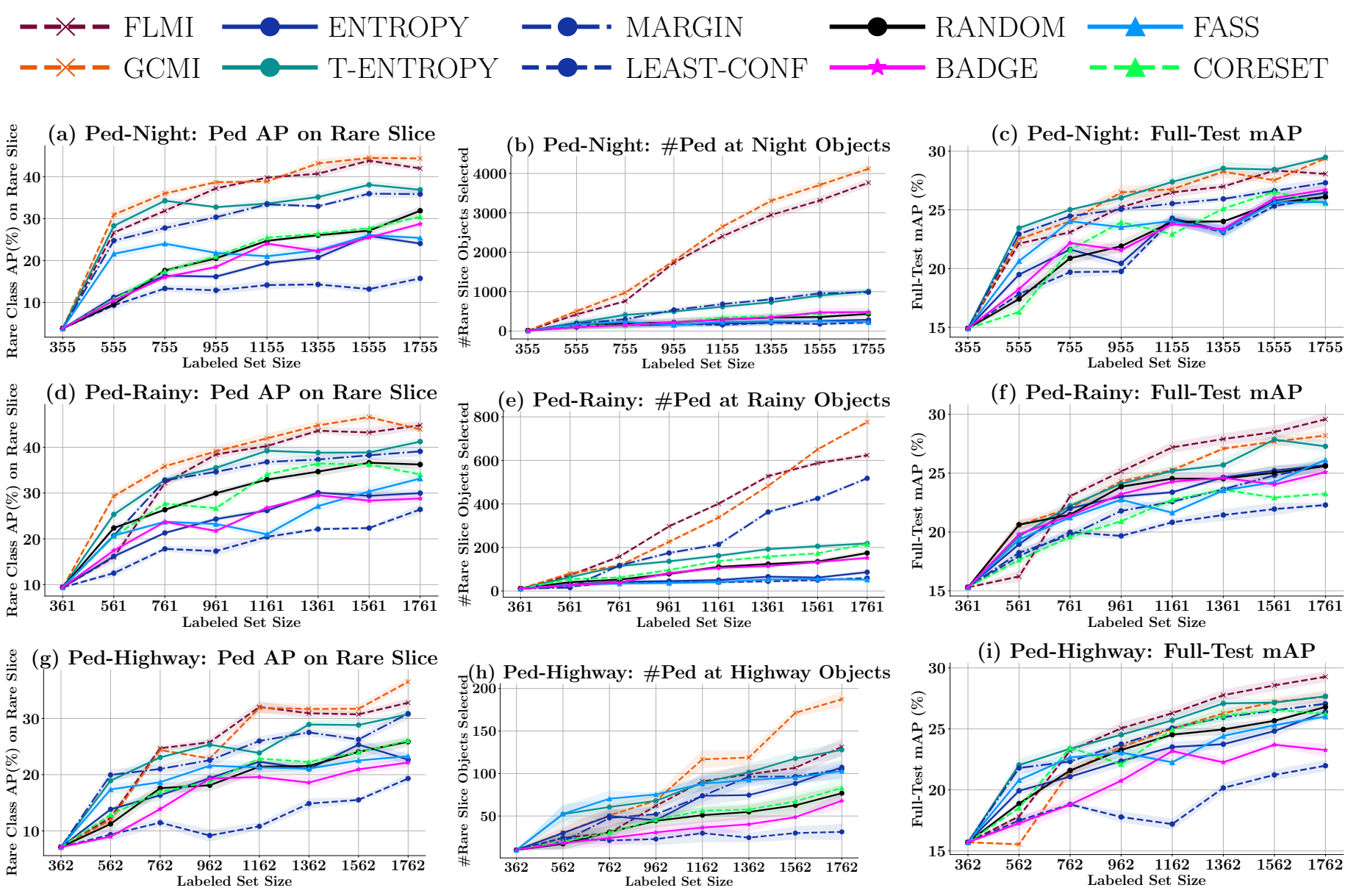

Fig. 8: AL with Pedestrian (**Ped**) at Nighttime (top row), Pedestrian in Rainy Weather (middle row), and Pedestrian on a Highway (bottom row) rare slices on BDD100K. Left side plots (a,d,g) show the AP of the rare class on the rare slice of data, center plots (b,e,h) show the number of objects selected that belong to the rare slice, and right side plots (c,f,i) show the mAP on the full test set of BDD100K. We observe that the SMI functions (FLMI, GCMI) outperform other baselines by $\approx 5\% - 10\%$ AP of the pedestrian class on the rare slice.

**Motorcycle *or* Bicycle at Night rare slice results:** We show the results for the 'motorcycle at night' and 'bicycle at night' rare slices in Fig. 7(top and middle row). We observe that the TALISMAN outperforms other baselines by $\approx 5\% - 14\%$ AP of the rare class on the rare slice (see Fig. 7 (a,d)), and by $\approx 2\%{-}4\%$ (see Fig. 7 (c,f)) in terms of mAP on the full test set. The gain in AP of the rare class and mAP increases in the later rounds of active learning, since the embedding representation of the model improves. Specifically, GCMI outperforms all methods since it models query-relevance well by selecting many rare class objects that belong to the rare slice (see Fig. 7(b,e)). In Fig. 9, we qualitatively show an example false negative motorcycle at night fixed using TALISMAN.

**Motorcycle *and* Bicycle at Night rare slice results:** We show the results for a scenario with multiple rare slices: 'motorcycle and bicycle at night' (Fig. 7(bottom row)). Importantly, we observe that Talisman selects more number of objects from both the rare slices in comparison to existing methods (see Fig. 7(h,i)). This is critical in real-world scenarios, since there are often cases with multiple co-occuring rare slices.

**Pedestrian at Night *or* Rainy *or* Highway rare slice results:** To study the robustness of Talisman in diverse real-world scenarios, we evaluate its performance for the 'pedestrian' rare class on multiple attributes - 1) 'night', 2) 'rainy', and 3) 'highway' (see Fig. 8). We observe consistent performance of both the Talisman instantiated functions (Gcmi and Flmi) across all scenarios. Concretely, we show that our framework can robustly find more pedestrians than any other baseline *across all rare slices* (see Fig. 8(b,e,h)), which leads to a performance gain of $\approx 5\% - 10\%$ AP over existing baselines for the pedestrian class on the rare slice. This reinforces the need for a framework like Talisman for improving the performance of object detectors on such rare slices.

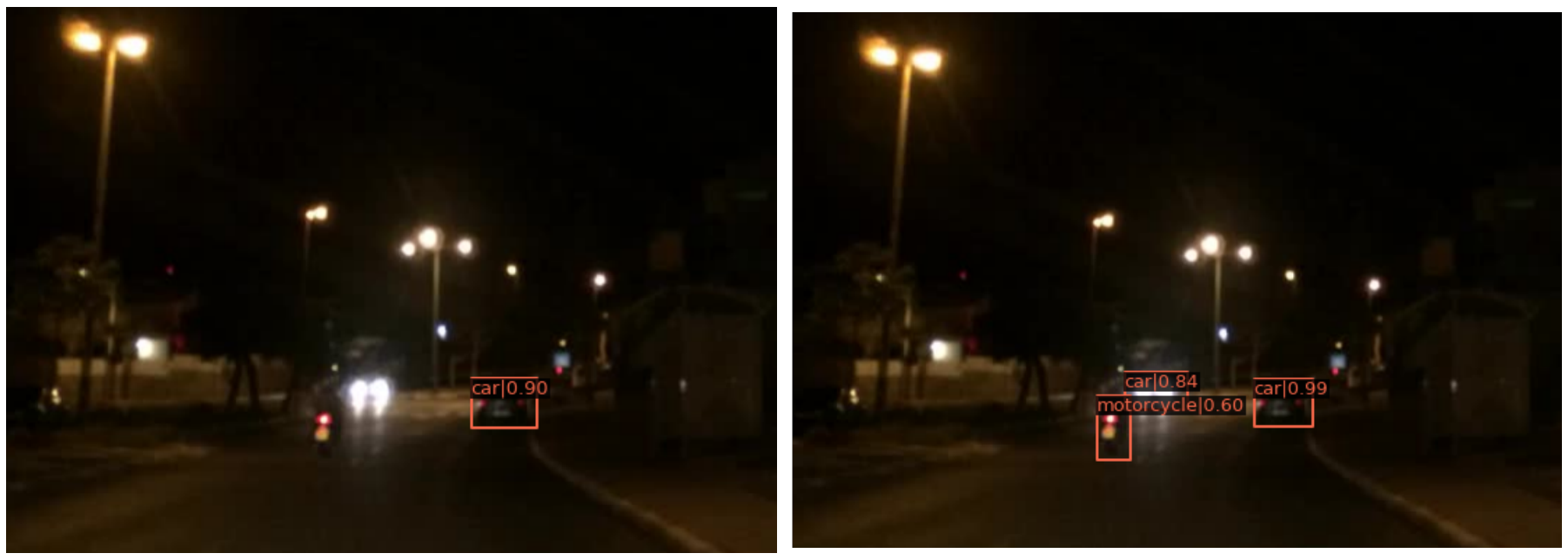

Fig. 9: A false negative motorcycle at night (left) fixed to a true positive detection (right) using Talisman.

## 6 Conclusion

In this paper, we present a targeted active learning framework Talisman that enables improving the performance of object detection models on rare classes and slices. We showed the utility of our framework across a variety of real-world scenarios with one or more rare classes and slices on the PASCAL VOC07+12 and BDD100K driving dataset, and observe a $\approx 5\% - 14\%$ gain compared to the existing baselines. Moreover, Talisman can select objects belonging to *multiple* co-occuring rare slices and simultaneously improve their performance, which is critical for modern object detectors. The main limitation of our work is the requirement of a reasonable feature embedding for computing similarity.

**Acknowledgments.** This work is supported by the National Science Foundation under Grant No. IIS-2106937, a gift from Google and Adobe, an Amazon Research Award, and an Adobe Data Science Award.

# Appendix

## A Summary of Notations

We summarize the notations used in this paper in Tab. 2.

## B Additional Experiments and Details

We provide the exact number of images in the initial labeled set $\mathcal{L}$ and the query set $\mathcal{Q}$ for each rare slice experiment in Tab. 3. Further, in Fig. 10 and Fig. 11 we qualitatively show how TALISMAN is able to fix a false negative motorcycle failure case. Lastly, in Fig. 12 we show the effectiveness of TALISMAN in another rare classes scenario on the BDD100K dataset. We observe that SMI functions used in TALISMAN outperform the baselines by $\approx 8\% - 14\%$ AP of the rare class and by $\approx 3\% - 4\%$ in terms of the mAP.

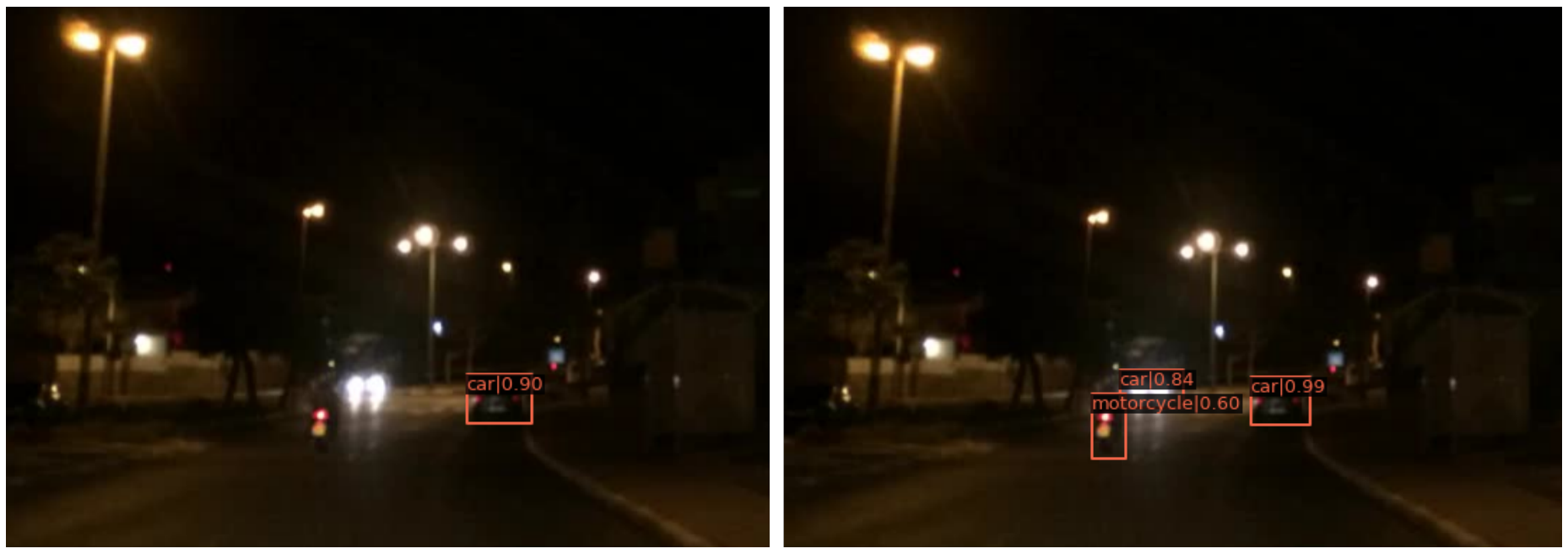

Fig. 10: A false negative motorcycle at night on the road (left) fixed to a true positive detection (right) using TALISMAN.

| Topic | Notation | Explanation |
|---|---|---|
| TALISMAN (Sec. 4) | $\mathcal{U}$ | Unlabeled set of $\|\mathcal{U}\|$ instances |
| | $\mathcal{A}$ | A subset of $\mathcal{U}$ |
| | $S_{ij}$ | Similarity between any two data points $i$ and $j$ |
| | $f$ | A submodular functions |
| | $\mathcal{L}$ | Labeled set of data points |
| | $\mathcal{Q}$ | Query set |
| | $\mathcal{M}$ | Object detection model |
| | $B$ | Active learning selection budget |
| | $\mathcal{I}_q$ | A single query image in the query set $\mathcal{Q}$ |
| | $\mathcal{I}_u$ | A single unlabeled image in the unlabeled set $\mathcal{U}$ |
| | $T$ | Number of region of interests (RoIs) in a single query image $\mathcal{I}_q$ |
| | $P$ | Number of proposals in a single unlabeled image $\mathcal{I}_u$ |
| | $\mathcal{E}_q$ | Embedding of $T$ RoIs in $\mathcal{I}_q$, $\mathcal{E}_q \in \mathbb{R}^{T \times D}$ |
| | $\mathcal{E}_u$ | Embedding of $P$ proposals in $\mathcal{I}_u$, $\mathcal{E}_u \in \mathbb{R}^{P \times D}$ |
| | $D$ | Dimensionality of each feature vector representing a RoI/proposal |
| | $F_\theta$ | Feature extraction module with parameters $\theta$ |
| | $\mathcal{X}_{qu}$ | Intermediate score map obtained by computing cosine similarity between $\mathcal{E}_q$ and $\mathcal{E}_u$, $\mathcal{X}_{qu} \in \mathbb{R}^{T \times P}$ |
| TALISMAN for Rare Classes (Sec. 5.3) | $\mathcal{C}_i^{\mathcal{L}}$ | Number of objects in $\mathcal{L}$ that belong to the rare (infrequent) class $i$. The total number of rare objects in $\mathcal{L}$ is $\|\mathcal{C}^{\mathcal{L}}\|$ |
| | $\mathcal{B}_j^{\mathcal{L}}$ | Number of objects in $\mathcal{L}$ that belong to the frequent class $j$. The total number of rare objects in $\mathcal{L}$ is $\|\mathcal{B}^{\mathcal{L}}\|$ |
| | $\rho$ | Imbalance ratio between $\mathcal{C}_i^{\mathcal{L}}$ and $\mathcal{B}_j^{\mathcal{L}}$ |
| TALISMAN for Rare Slices (Sec. 5.4) | $\|\mathcal{O}_c^A\|$ | Total number of objects in $\mathcal{L}$ that belong to class $c$ and *have* the attribute $A$ |
| | $\|\mathcal{O}_c^{\tilde{\mathbb{A}}}\|$ | Total number of objects in $\mathcal{L}$ that belong to class $c$ and *do not have* the attribute $A$ |
| | $\rho$ | Imbalance ratio between $\|\mathcal{O}_c^A\|$ and $\|\mathcal{O}_c^{\tilde{\mathbb{A}}}\|$ |

Table 2: Summary of notations used throughout this paper

Table 3: Number of data points in the initial labeled set $\mathcal{L}$ and query set $\mathcal{Q}$ for different rare slice experiments in Sec. 5.4

| **Rare Slice** | Initial $\mathcal{L}$ Size (# Images) | Query Set $\mathcal{Q}$ Size (# Images) |
|---|---|---|
| Motorcycle at Nighttime | 363 | 5 |
| Bicycle at Nighttime | 348 | 5 |
| Pedestrian at Nighttime | 355 | 3 |
| Pedestrian in Rainy Weather | 361 | 5 |
| Pedestrian on a Highway | 362 | 5 |

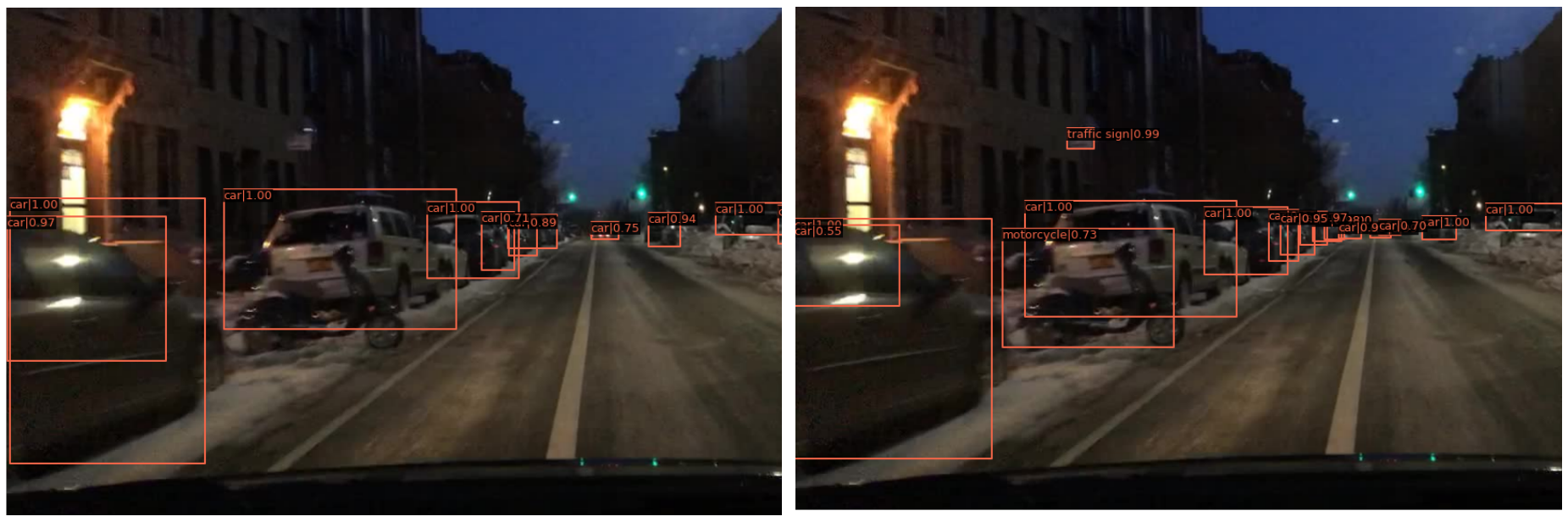

Fig. 11: A false negative motorcycle at night parked on the road (left) fixed to a true positive detection (right) using Talisman.

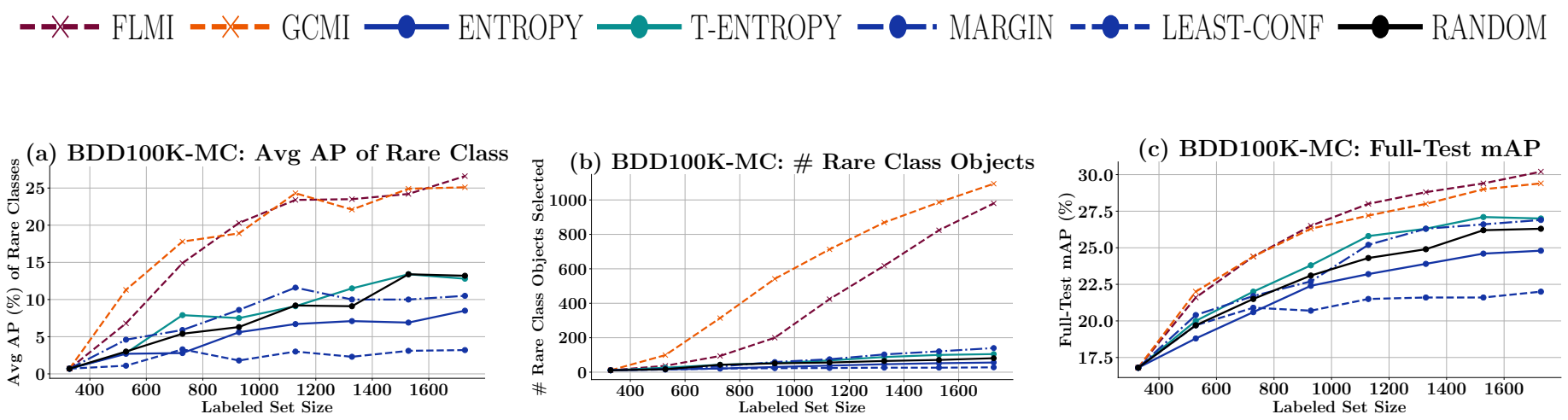

Fig. 12: Active Learning with Motorcycle (**MC**) rare class on BDD100K. Left side plot (a) shows the AP of the rare class, center plot (b) show the number of objects selected that belong to the rare class, and right side plots (c) show the mAP on the full test set of BDD100K. We observe that the SMI functions (Flmi, Gcmi) outperform other baselines by $\approx 8\% - 14\%$ AP of the rare class.